\documentclass[conference]{IEEEtran}
\IEEEoverridecommandlockouts
\usepackage{cite}
\usepackage[british]{babel}
\usepackage[utf8]{vietnam}
\usepackage{amsmath,amssymb,amsfonts}
\usepackage{algorithmic}
\usepackage{graphicx}
\usepackage{textcomp}
\usepackage{xcolor}
\def\BibTeX{{\rm B\kern-.05em{\sc i\kern-.025em b}\kern-.08em
    T\kern-.1667em\lower.7ex\hbox{E}\kern-.125emX}}
\begin{document}
\selectlanguage{british}
\title{VAIS Hate Speech Detection System: \\A Deep Learning based Approach \\ for System Combination
% \\
% {\footnotesize \textsuperscript{*}VAIS: Vietnam Artificial Intelligence System}
}

\author{\IEEEauthorblockN{1\textsuperscript{st} Thai Binh Nguyen}
\IEEEauthorblockA{\textit{Vietnam Artificial Intelligence System} \\
\textit{Hanoi University of Science and Technology}\\
Hanoi, Vietnam \\
binhnguyen@vais.vn}
\and
\IEEEauthorblockN{2\textsuperscript{nd} Quang Minh Nguyen}
\IEEEauthorblockA{\textit{Vietnam Artificial Intelligence System} \\
% \textit{FPT University}\\
Hanoi, Vietnam \\
minhnq@vais.vn}

\and
\IEEEauthorblockN{3\textsuperscript{rd} Thu Hien Nguyen}
\IEEEauthorblockA{\textit{Thai Nguyen University of Education} \\
Thai Nguyen, Vietnam \\
nguyenthuhien@dhsptn.edu.vn}

\and
\IEEEauthorblockN{4\textsuperscript{rd} Ngoc Phuong Pham}
\IEEEauthorblockA{\textit{Vietnam Artificial Intelligence System} \\
\textit{Thai Nguyen University} \\
Thai Nguyen, Vietnam \\
phuongpn@tnu.edu.vn}

\and
\IEEEauthorblockN{5\textsuperscript{rd} The Loc Nguyen}
\IEEEauthorblockA{
\textit{Vietnam Artificial Intelligence System} \\
\textit{Hanoi University of Mining and Geology} \\
Hanoi, Vietnam \\
locnguyen@vais.vn}

\and
\IEEEauthorblockN{6\textsuperscript{rd} Quoc Truong Do}
\IEEEauthorblockA{\textit{Vietnam Artificial Intelligence System} \\
Hanoi, Vietnam \\
truongdo@vais.vn}

% \and
% \and
% \and
}

\maketitle

% \begin{abstract}
% This document is a model and instructions for \LaTeX.
% This and the IEEEtran.cls file define the components of your paper [title, text, heads, etc.]. *CRITICAL: Do Not Use Symbols, Special Characters, Footnotes, 
% or Math in Paper Title or Abstract.
% \end{abstract}

\begin{abstract}
Nowadays, Social network sites (SNSs) such as Facebook, Twitter are common places where people show their opinions, sentiments and share information with others. However, some people use SNSs to post abuse and harassment threats in order to prevent other SNSs users from expressing themselves as well as seeking different opinions. To deal with this problem, SNSs have to use a lot of resources including people to clean the aforementioned content. In this paper, we propose a supervised learning model based on the ensemble method to solve the problem of detecting hate content on SNSs in order to make conversations on SNSs more effective. Our proposed model got the first place for public dashboard with 0.730 F1 macro-score and the third place with 0.584 F1 macro-score for private dashboard at the sixth international workshop on Vietnamese Language and Speech Processing 2019.
\end{abstract}


\section{Introduction}
Currently, social networks are so popular. Some of the biggest ones include Facebook, Twitter, Youtube,... with extremely number of users. Thus, controlling content of those platforms is essential. For years, social media companies such as Twitter, Facebook, and YouTube have been investing hundreds of millions euros on this task \cite{gamback2017using, TheGuardian2016}. However, their effort is not enough since such efforts are primarily based on manual moderation to identify and delete offensive materials. The process is labour intensive, time consuming, and not sustainable or scalable in reality \cite{chen2012detecting, gamback2017using, waseem2016hateful}.

In the sixth international workshop on Vietnamese Language and Speech Processing (VLSP 2019), the Hate Speech Detection (HSD) task is proposed as one of the shared-tasks to handle the problem related to controlling content in SNSs. HSD is required to build a multi-class classification model that is capable of classifying an item to one of 3 classes (\textit{hate}, \textit{offensive}, \textit{clean}). Hate speech (\textit{hate}): an item is identified as hate speech if it (1) targets individuals or groups on the basis of their characteristics; (2) demonstrates a clear intention to incite harm, or to promote hatred; (3) may or may not use offensive or profane words. Offensive but not hate speech (\textit{offensive}): an item (posts/comments) may contain offensive words but it does not target individuals or groups on the basis of their characteristics. Neither offensive nor hate speech (\textit{clean}): normal item, it does not contain offensive language or hate speech. 

The term `hate speech' was formally defined as `any communication that disparages a person or a group on the basis of some characteristics (to be referred to as types of hate or hate classes) such as race, colour, ethnicity, gender, sexual orientation, nationality, religion, or other characteristics' \cite{nockleby1994hate}. Many researches have been conducted in recent years to develop automatic methods for hate speech detection in the social media domain. These typically employ semantic content analysis techniques built on Natural Language Processing (NLP) and Machine Learning (ML) methods. The task typically involves classifying textual content into non-hate or hateful. This HSD task is much more difficult when it requires classify text in three classes, with \textit{hate} and \textit{offensive} class quite hard to classify even with humans. 

\begin{figure*}[t]
  \centering
  \includegraphics[width=0.9\linewidth]{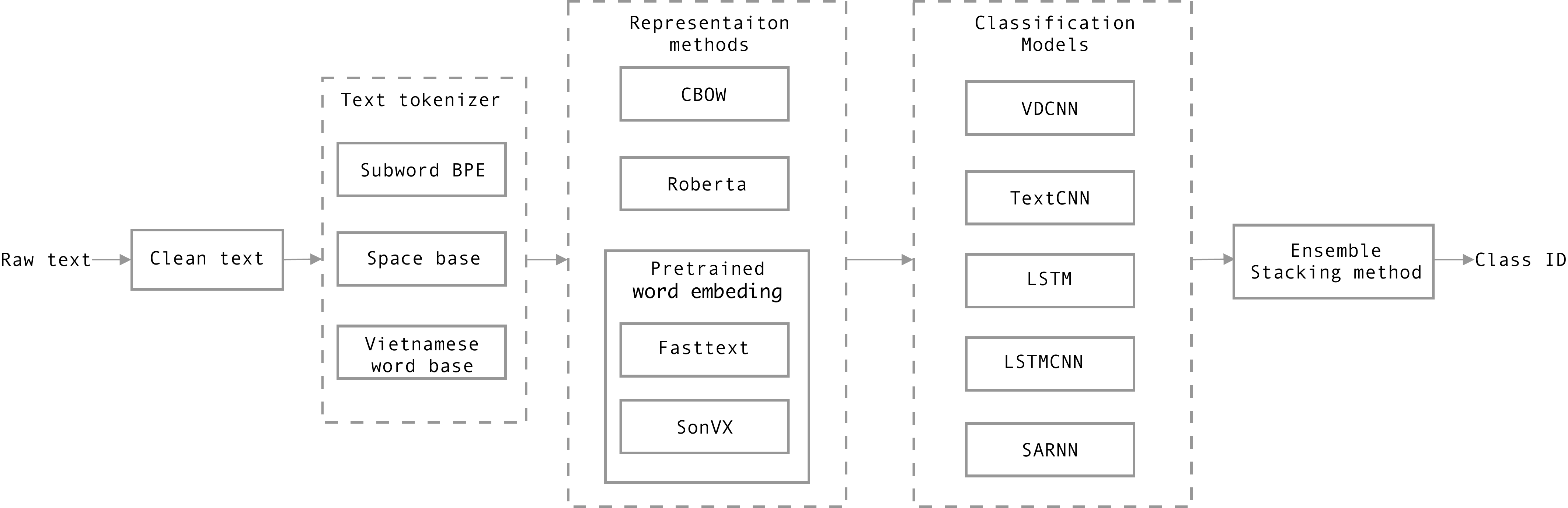}
  \caption{Hate Speech Detection System Overview}
  \label{fig:system-overview}
\end{figure*}

\begin{figure}[t]
  \centering
  \includegraphics[width=0.7\linewidth]{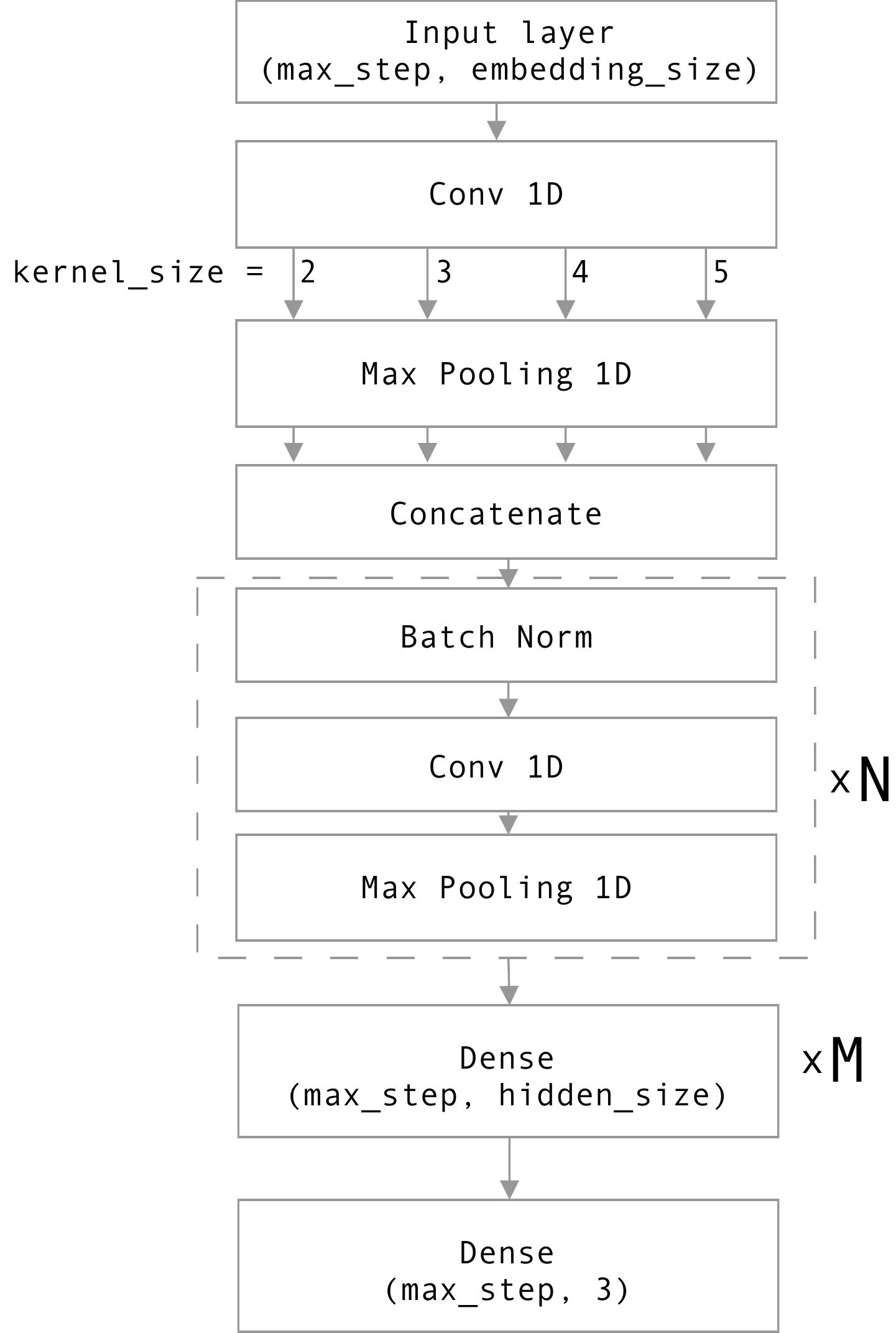}
  \caption{TextCNN model architecture}
  \label{fig:textcnn}
\end{figure}

In this paper, we propose a method to handle this HSD problem. Our system combines multiple text representations and models architecture in order to make diverse predictions. The system is heavily based on the ensemble method. The next section will present detail of our system including data preparation (how we clean text and build text representation), architecture of the model using in the system, and how we combine them together. The third section is our experiment and result report in HSD shared-task VLSP 2019. The final section is our conclusion with advantages and disadvantages of the system following by our perspective. 

\section{System description}
In this section, we present the system architecture. It includes how we pre-process text, what types of text representation we use and models used in our system. In the end, we combine model results by using an ensemble technique.

\subsection{System overview}
The fundamental idea of this system is how to make a system that has the diversity of viewing an input. That because of the variety of the meaning in Vietnamese language especially with the acronym, teen code type. To make this diversity, after cleaning raw text input, we use multiple types of word tokenizers. Each one of these tokenizers, we combine with some types of representation methods, including word to vector methods such as continuous bag of words \cite{mikolov2013efficient}, pre-trained embedding as fasttext (trained on Wiki Vietnamese language) \cite{joulin2016fasttext} and sonvx (trained on Vietnamese newspaper) \cite{word2vecvn_2016}. Each sentence has a set of words corresponding to a set of word vectors, and that set of word vectors is a representation of a sentence. We also make a sentence embedding by using RoBERTa architecture \cite{liu2019roberta}. CBOW and RoBERTa models trained on text from some resources including VLSP 2016 Sentiment Analysis, VLSP 2018 Sentiment Analysis, VLSP 2019 HSD and text crawled from Facebook. After having sentence representation, we use some classification models to classify input sentences. Those models will be described in detail in the section \ref{model-archi}. With the multiply output results, we will use an ensemble method to combine them and output the final result. Ensemble method we use here is Stacking method will be introduced in the section \ref{ensemble-section}.

\subsection{Data pre-processing}
% Steps process text from HSD dataset
% Data source include using in this system
% Tokenizer type (word base, Vietnamese word tokenizer, subword bpe)
% Text and sentence to vector
Content in the dataset that provided in this HSD task is very diverse. Words having the same meaning were written in various types (teen code, non tone, emojis,..) depending on the style of users. Dataset was crawled from various sources with multiple text encodes. In order to make it easy for training, all types of encoding need to be unified. This cleaning module will be used in two processes: cleaning data before training and cleaning input in inferring phase. Following is the data processing steps that we use:

\begin{figure}[t]
  \centering
  \includegraphics[width=0.7\linewidth]{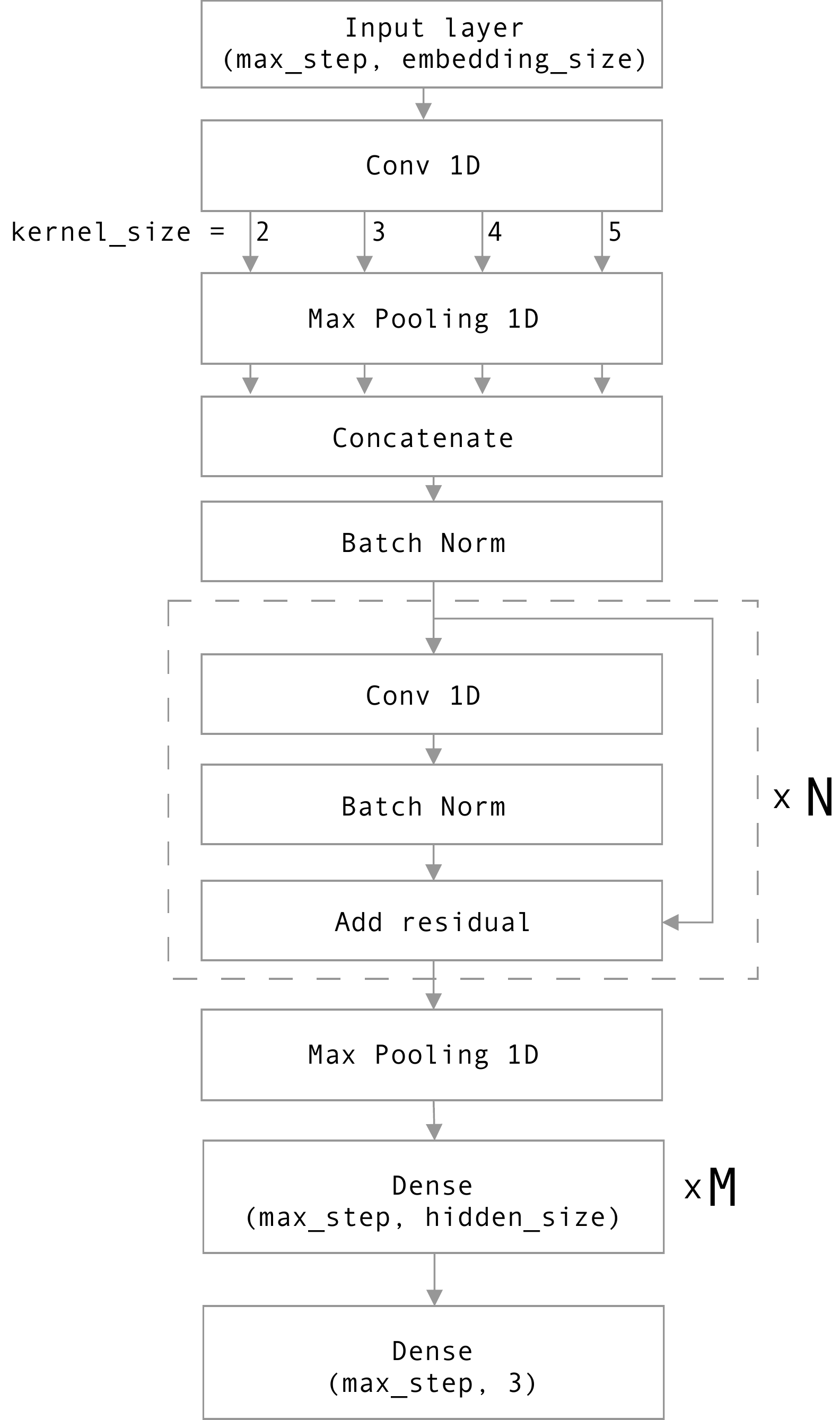}
  \caption{VDCNN model architecture}
  \label{fig:vdcnn}
\end{figure}

\begin{itemize}
    \item \textbf{Step 1}: Format encoding. Vietnamese has many accents, intonations with different Unicode typing programs which may have different outputs with the same typing type. To make it unified, we build a library named \textit{visen}\footnote{https://github.com/nguyenvulebinh/visen}. For example, the input "thíêt kê\'" will be normalized to "thiết kế" as the output.
    
    \item \textbf{Step 2}: In social networks, people show their feelings a lot by emojis. Emoticon is often a special Unicode character, but sometimes, it is combined by multiple normal characters like `: ( = ]'. We make a dictionary mapping this emoji (combined by some characters) to a single Unicode character like other emojis to make it unified.

    \item \textbf{Step 3}: Remove unseen characters. For human, unseen character is invisible but for a computer, it makes the model harder to process and inserts space between words, punctuation and emoji. This step aims at reducing the number of words in the dictionary which is important task, especially with low dataset resources like this HSD task.
    
    \item \textbf{Step 4}: With model requiring Vietnamese word segmentation as the input, we use \cite{vu-etal-2018-vncorenlp,nguyen-etal-2017-word} to tokenize the input text.
    
    \item \textbf{Step 5}: Make all string lower. We experimented and found that lower-case or upper-case are not a significant impact on the result, but with lower characters, the number of words in the dictionary is reduced.    
    
\end{itemize}

RoBERTa proposed in \cite{liu2019roberta} an optimized method for pretraining self-supervised NLP systems. In our system, we use RoBERTa not only to make sentence representation but also to augment data. With mask mechanism, we replace a word in the input sentence with another word that RoBERTa model proposes. To reduce the impact of replacement word, the chosen words are all common words that appear in almost three classes of the dataset. For example, with input \textit{`nhổn làm gắt vl'}, we can augment to other outputs: \textit{`vl làm gắt qá'}, \textit{`còn làm vl vậy'}, \textit{`vl làm đỉnh vl'} or \textit{`thanh chút gắt vl'}.

\selectlanguage{british}
\begin{table*}[ht]
\small
\centering
\begin{tabular}{|c|c|c|c|c|c|}
    \hline  
    Embedding & VDCNN & TextCNN & LSTM & LSTMCNN & SARNN \\
    \hline
    comment & 0.6812 & 0.6743 & 0.6612 & 0.7012 & 0.7056 \\
    \hline
    comment\_bpe & 0.6643 & 0.6665 & 0.6514 & 0.6918 & 0.6901  \\
    \hline
    comment\_tokenize & 0.7098 & 0.7143 & 0.6832 & 0.7123 & \textbf{0.7167} \\
    \hline
    fasttext & 0.6954 & 0.7123 & 0.6812 & 0.7012 & 0.7012 \\
    \hline
    roberta & 0.6734 & 0.6636 & 0.6345 & 0.6704 & 0.6566 \\
    \hline
    sonvx\_wiki & 0.6534 & 0.6624 & 0.6456 & 0.6745 & 0.6316 \\
    \hline
    sonvx\_baomoi\_w2 & 0.6612 & 0.6712 & 0.6549 & 0.6822 & 0.6513 \\
    \hline
    sonvx\_baomoi\_w5 & 0.6656 & 0.6645 & 0.6601 & 0.6756 & 0.6647 \\
    \hline
\end{tabular}
\caption{F1\_macro score of different model}
\label{f1_macro}
\end{table*}

\subsection{Models architecture}\label{model-archi}

Social comment dataset has high variety, the core idea is using multiple model architectures to handle data in many viewpoints. In our system, we use five different model architectures combining many types of CNN, and RNN. Each model will use some types of word embedding or handle directly sentence embedding to achieve the best general result. Source code of five models is extended from the GitHub repository\footnote{https://github.com/petrpan26/Aivivn\_1}

The first model is TextCNN (figure \ref{fig:textcnn}) proposed in \cite{kim-2014-convolutional}. It only contains CNN blocks following by some Dense layers. The output of multiple CNN blocks with different kernel sizes is connected to each other.

The second model is VDCNN (figure \ref{fig:vdcnn}) inspired by the research in \cite{Conneau_2017}. Like the TextCNN model, it contains multiple CNN blocks. The addition in this model is its residual connection.

\begin{figure}[t]
  \centering
  \includegraphics[width=0.5\linewidth]{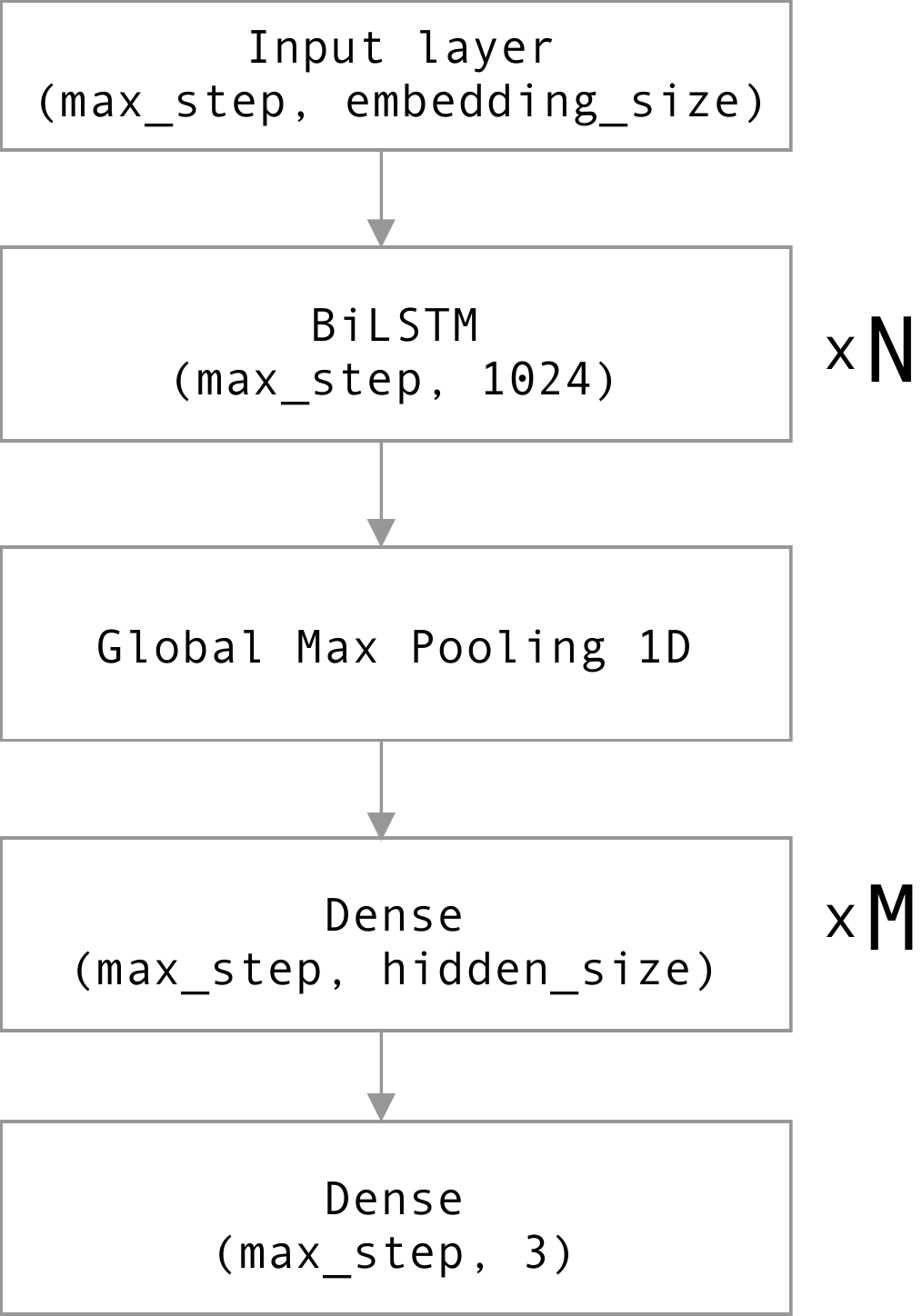}
  \caption{LSTM model architecture}
  \label{fig:lstm}
\end{figure}

The third model is a simple LSTM bidirectional model (figure \ref{fig:lstm}). It contains multiple LSTM bidirectional blocks stacked to each other.

\begin{figure}[t]
  \centering
  \includegraphics[width=0.7\linewidth]{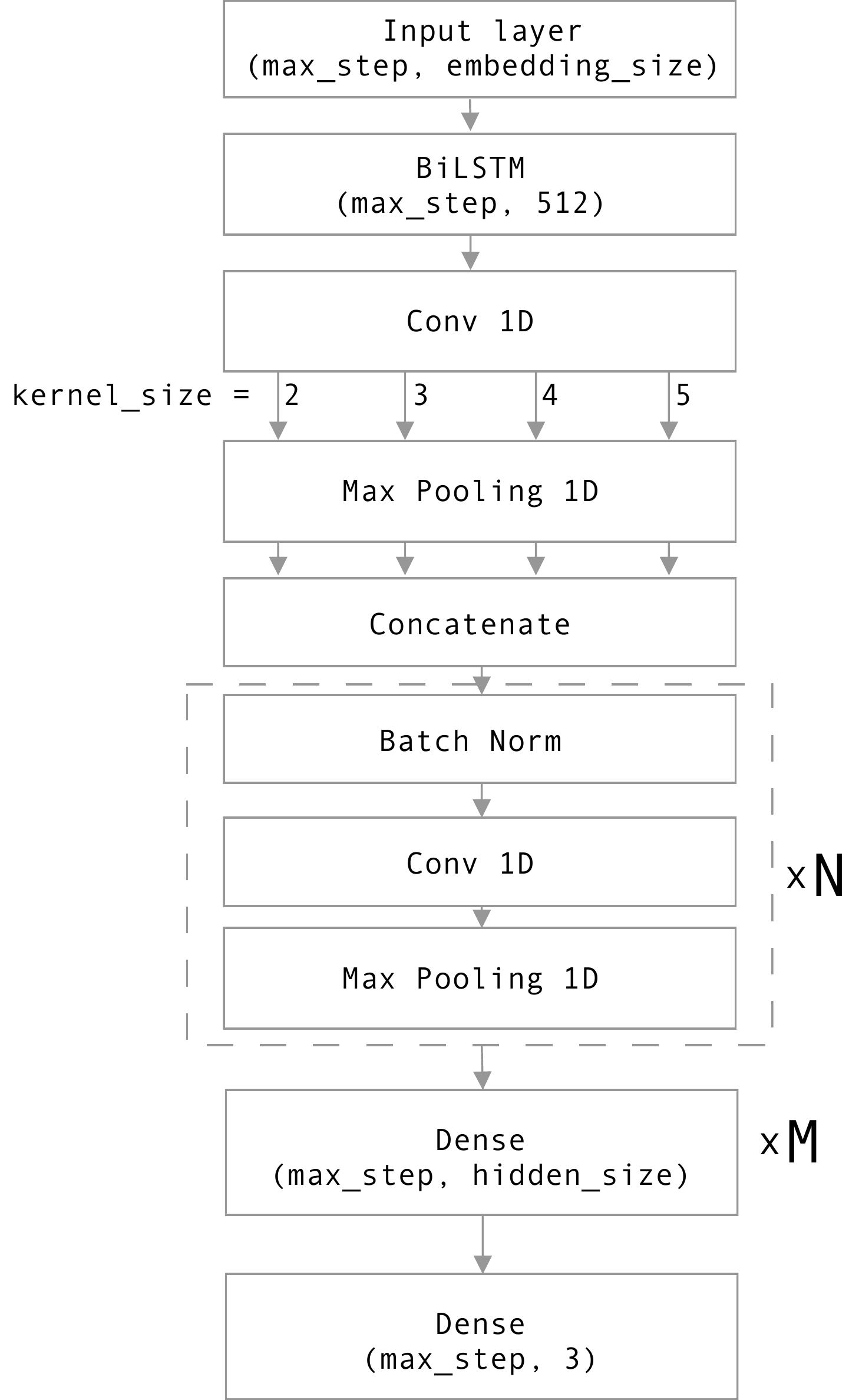}
  \caption{LSTMCNN model architecture}
  \label{fig:lstmcnn}
\end{figure}

The fourth model is LSTMCNN (figure \ref{fig:lstmcnn}). Before going through CNN blocks, series of word embedding will be transformed by LSTM bidirectional block.

\begin{figure}[t]
  \centering
  \includegraphics[width=0.5\linewidth]{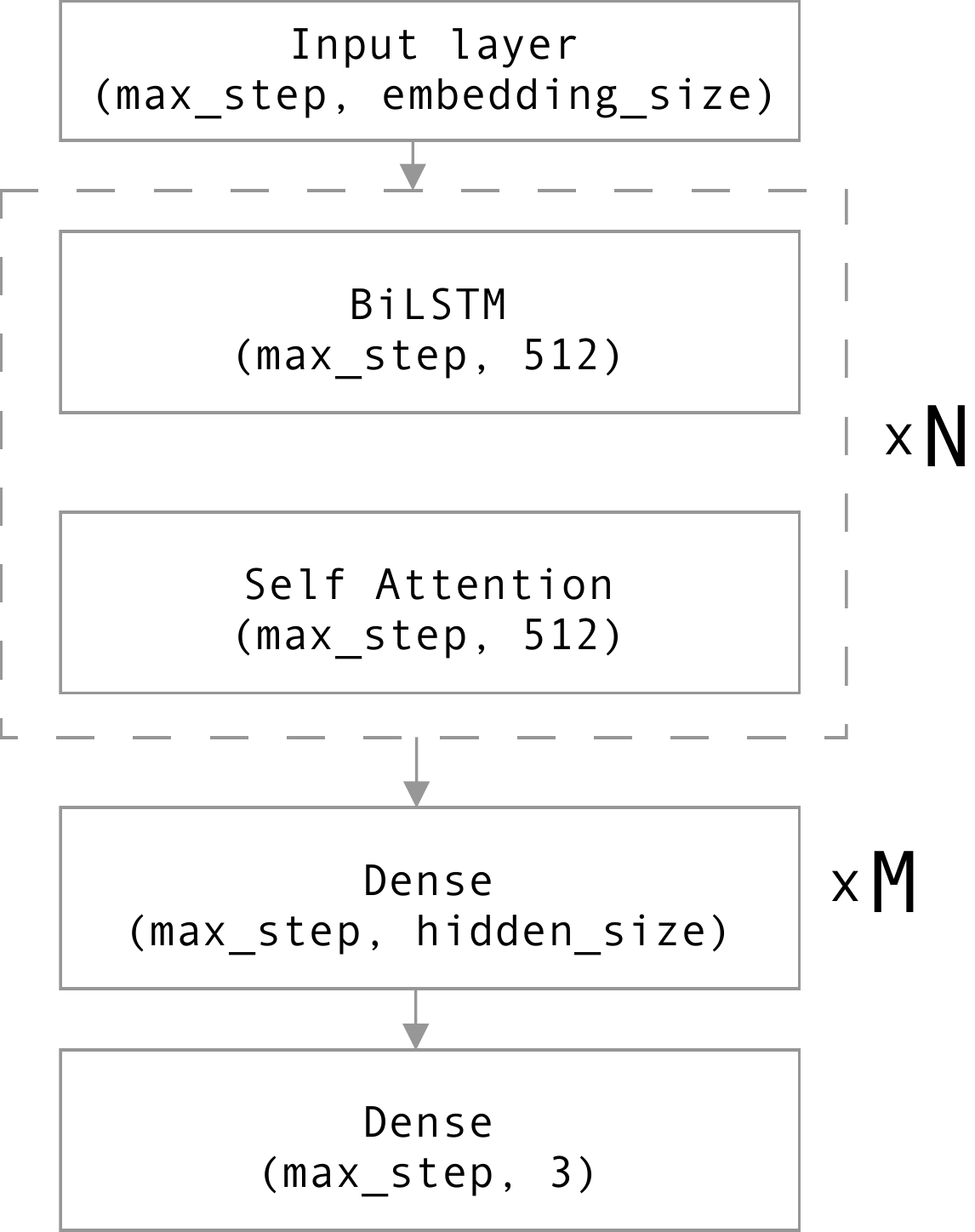}
  \caption{SARNN model architecture}
  \label{fig:sarnn}
\end{figure}

The final model is the system named SARNN (figure \ref{fig:sarnn}). It adds an attention block between LTSM blocks.

\subsection{Ensemble method}\label{ensemble-section}

Ensemble methods is a machine learning technique that combines several base models in order to produce one optimal predictive model. Have the main three types of ensemble methods including Bagging, Boosting and Stacking. In this system, we use the Stacking method. In this method, the output of each model is not only class id but also the probability of each class in the set of three classes. This probability will become a feature for the ensemble model. The stacking ensemble model here is a simple full-connection model with input is all of probability that output from sub-model. The output is the probability of each class. 

\section{Experiment}
% Express about the data (imbalance, ....)
The dataset in this HSD task is really imbalance. \textit{Clean} class dominates with 91.5\%, \textit{offensive} class takes 5\% and the rest belongs to \textit{hate} class with 3.5\%. To make model being able to learn with this imbalance data, we inject class weight to the loss function with the corresponding ratio (\textit{clean}, \textit{offensive}, \textit{hate}) is $(0.09, 0.95, 0.96)$. Formular \ref{loss_weight} is the loss function apply for all models in our system. $w_i$ is the class weight, $y_i$ is the ground truth and $\hat{y}_i$ is the output of the model. If the class weight is not set, we find that model cannot adjust parameters. The model tends to output all \textit{clean} classes.

\begin{align}
    J = -\frac{1}{N}(\sum^N_{i=1}w_i*y_i*log(\hat{y}_i))
\label{loss_weight}
\end{align}

We experiment 8 types of embedding in total:
\begin{itemize}
    \item \textbf{comment}: CBOW embedding training in all dataset comment, each word is splited by space. Embedding size is 200.
    \item \textbf{comment\_bpe}: CBOW embedding training in all dataset comment, each word is splited by subword bpe\footnote{https://github.com/rsennrich/subword-nmt}.  Embedding size is 200.
    \item \textbf{comment\_tokenize}: CBOW embedding training in all dataset comment, each word is splited by space. Before split by space, word is concatenated by using \cite{vu-etal-2018-vncorenlp, ng18,nguyen-etal-2017-word}.  Embedding size is 200.
    \item \textbf{roberta}: sentence embedding training in all dataset comment, training by using RoBERTa architecture.  Embedding size is 256.
    \item \textbf{fasttext, sonvx*} is all pre-trained word embedding in general domain. Before mapping word to vector, word is concatenated by using \cite{vu-etal-2018-vncorenlp, ng18,nguyen-etal-2017-word}. Embedding size of fasttext is 300. (sonvx\_wiki, sonvx\_baomoi\_w2, sonvx\_baomoi\_w5) have embedding size corresponding is (400, 300, 400).
    
\end{itemize}

% Train valid split
In our experiment, the dataset is split into two-part: \textit{train} set and \textit{dev} set with the corresponding ratio $(0.9, 0.1)$. Two subsets have the same imbalance ratio like the root set. For each combination of model and word embedding, we train model in \textit{train} set until it achieve the best result of loss score in the \textit{dev} set. The table \ref{f1_macro} shows the best result of each combination on the f1\_macro score. 

For each model having the best fit on the \textit{dev} set, we export the probability distribution of classes for each sample in the \textit{dev} set. In this case, we only use the result of model that has f1\_macro score that larger than 0.67. The probability distribution of classes is then used as feature to input into a dense model with only one hidden layer (size 128). The training process of the ensemble model is done on samples of the \textit{dev} set. The best fit result is 0.7356. The final result submitted in public leaderboard is 0.73019 and in private leaderboard is 0.58455. It is quite different in bad way. That maybe is the result of the model too overfit on \textit{train} set tuning on public \textit{test} set. 

% Analysis error in valid part
Statistics of the final result on the \textit{dev} set shows that almost cases have wrong prediction from \textit{offensive} and \textit{hate} class to \textit{clean} class belong to samples containing the word `vl'. (62\% in the \textit{offensive} class and 48\% in the \textit{hate} class). It means that model overfit the word `vl' to the \textit{clean} class. This makes sense because `vl' appears too much in the \textit{clean} class dataset.

In case the model predicts wrong from the \textit{clean} class to the \textit{offensive} class and the \textit{hate} class, the model tends to decide case having sensitive words to be wrong class. The class \textit{offensive} and the \textit{hate} are quite difficult to distinguish even with human. 

\section{Conclusion}

In this study, we experiment the combination of multiple embedding types and multiple model architecture to solve a part of the problem Hate Speech Detection with a signification good classification results. Our system heavily based on the ensemble technique so the weakness of the system is slow processing speed. But in fact, it is not big trouble with this HSD problem when human usually involve handling directly in the before. 

HSD is a hard problem even with human. In order to improve classification quality, in the future, we need to collect more data especially social networks content. This will make building text representation more correct and help model easier to classify.

\selectlanguage{british}
\bibliography{tailieu}

\end{document}